\documentclass[letterpaper, 10 pt, conference]{ieeeconf}  

\IEEEoverridecommandlockouts                              

\usepackage{graphics} 
\usepackage{epsfig} 
\usepackage{mathptmx} 
\usepackage{times} 
\usepackage{csquotes} 
\usepackage{amsmath} 
\usepackage{algorithm}
\usepackage{algorithmic}
\usepackage{tabularx}
\usepackage{makecell}
\usepackage{booktabs}
\usepackage{amssymb}  
\usepackage{graphicx}
\usepackage{xcolor}
\usepackage{comment}
\usepackage{subcaption}

\usepackage[bookmarks=false]{hyperref}

\usepackage{textcomp}
\usepackage{xcolor}
\usepackage{epsfig} 
\usepackage{mathptmx} 
\usepackage{times} 
\usepackage{amsmath} 
\usepackage{amssymb}  
\usepackage{cite}

\usepackage{array}
\usepackage{listings}
\title{\LARGE \bf
Instruct2Act: From Human Instruction to Actions Sequencing and Execution via Robot Action Network for Robotic Manipulation
}

\author{Archit Sharma, Dharmendra Sharma, John Rebeiro, Peeyush Thakur, Narendra Dhar, and Laxmidhar Behera\\
Indian Institute of Technology Mandi, India}

\begin{document}

\maketitle
\thispagestyle{empty}
\pagestyle{empty}

\begin{abstract}
Robots often struggle to follow free-form human instructions in real-world settings due to computational and sensing limitations. We address this gap with a lightweight, fully on-device pipeline that converts natural-language commands into reliable manipulation. Our approach has two stages: (i) the instruction to actions module (Instruct2Act), a compact BiLSTM with a multi-head-attention autoencoder that parses an instruction into an ordered sequence of atomic actions (e.g., reach, grasp, move, place); and (ii) the robot action network (RAN), which uses the dynamic adaptive trajectory radial network (DATRN) together with a vision-based environment analyzer (YOLOv8) to generate precise control trajectories for each sub-action. The entire system runs on a modest system with no cloud services. On our custom proprietary dataset, Instruct2Act attains 91.5\% sub-actions prediction accuracy while retaining a small footprint. Real-robot evaluations across four tasks (pick-place, pick-pour, wipe, and pick-give) yield an overall 90\% success; sub-action inference completes in $< 3.8\,\mathrm{s}$, with end-to-end executions in $30\text{--}60\,\mathrm{s}$ depending on task complexity. These results demonstrate that fine-grained instruction-to-action parsing, coupled with DATRN-based trajectory generation and vision-guided grounding, provides a practical path to deterministic, real-time manipulation in resource-constrained, single-camera settings.
\end{abstract}

\section{Introduction}

Robots are increasingly deployed in real-world environments to assist humans with complex tasks, particularly in industrial and healthcare settings where precise object manipulation is essential.  In healthcare, for instance, robots support clinical workflows by delivering medications, transporting instruments, cleaning surfaces, and assisting with patient care\cite{hosiptals}. Human instructions can be conveyed to robots via different modalities, including speech, text, gestures, and demonstrations. Among these, text-based instructions represent one of the simplest and most effective approaches \cite{Intro_RAL_1}. To interpret such instructions, robots rely on natural language processing (NLP), which enables them to understand and respond to human language. Unlike fixed menus or rigid programming, natural language provides flexibility and user-friendliness. It provides an intuitive way to describe complex tasks and is well-suited for everyday human environments \cite{NLP_robot_survey}, where instructions are often unstructured and context-dependent. This makes natural language the most practical option for smooth and effective communication between humans and robots.

By enabling robots to grasp and react to free-form natural language commands, recent approaches like large language models (LLMs) and vision language models (VLMs) have significantly advanced human-robot interaction \cite{4,7,8,VLM,GPT-4V(ision)}. To further enhance language understanding and perception, vision-language models (VLMs) integrate text and visual inputs to learn rich, multimodal representations. Building on this, recent vision-language action (VLA) pipelines map free-form instructions directly to object-aware perception and trajectory generation, allowing robots to follow commands in real time \cite{1,RobotGPT,openvla,tinyvla}. These advances make it possible for robots to perform context-aware manipulation, navigation, and multi-step tasks in open and dynamic environments.

\begin{figure}[H]
  \centering
  \includegraphics[width=0.90\columnwidth]{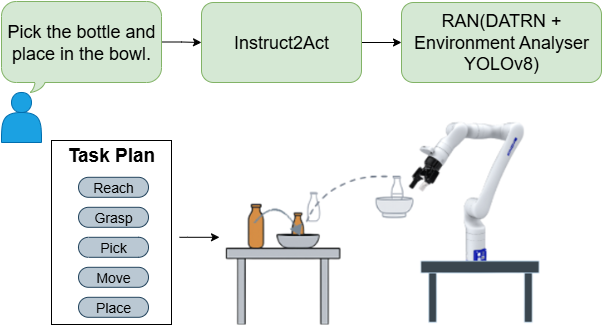}
  \caption{Overview of Instruct2Act and RAN.}
  \label{fig:front page workflow}
\end{figure}
However, implementing end-to-end VLA stacks in a camera setup that only uses eye-in-hand (wrist) configurations remains challenging. Relying on only an eye-in-hand camera in changing environments is brittle: the arm and gripper frequently occlude key objects, viewpoints shift rapidly, and limited global context induces pose and frame ambiguity, degrading manipulation reliability. Even lightweight VLA variants \cite{smolvla} often rely on wide-angle multi-camera setups or keep large multimodal encoders with long context windows. These designs introduce latency and increase maintenance costs in real-world environments. Such issues indicate practical deployment challenges rather than flaws in the VLA methods, which motivates the approach taken in this work.


To overcome these barriers, our primary objective is to develop a framework capable of executing sophisticated tasks without increasing the complexity of the system. In this paper, we developed a methodology that converts free-form natural language commands into robot actions and executes them entirely on-device. It is a two-stage pipeline that integrates (i) instruction-to-action prediction and (ii) robot action execution. 
We introduced Instruct2Act, a compact BiLSTM with a multi-head attention autoencoder, to decompose language commands into atomic sub-actions. This lightweight design aims to produce fine-grained action sequences with low computational overhead.

The prediction module is linked to a robot action network (RAN) that creates precise, adaptive motion trajectories for each sub-action, as illustrated in Fig.~\ref{fig:front page workflow}. RAN consists of a dynamic adaptive trajectory radial network (DATRN), which generates object-aware motion paths and adjusts them in real time using the integrated depth camera and proprioception of the manipulator. We compared our DATRN-based trajectory generation approach with the dynamic movement primitive (DMP) to overcome its limitations, such as iterative hyperparameter searches, phase/gain scheduling, and long fitting times \cite{Saveriano}. The entire pipeline operates with low computational overhead, demonstrating its suitability for resource-constrained environments. The reliability of this methodology was validated in controlled laboratory experiments and preliminary trials in real-world healthcare environments. This work demonstrates a practical pathway toward autonomous robots that can operate effectively in real-world settings without reliance on massive, off-board computational resources.


The practical \textbf{motivation} behind this work is to address the following challenges of robot manipulator execution:

\begin{itemize}
\item \textit{Accuracy and precision:} Provide fine-grained grounding of free-form natural language into clear, deterministic action plans that support precise and reliable robotic manipulation.

\item \textit{Robustness with eye-in-hand sensing:} Using a single wrist camera and maintaining stable perception for different source–different goal settings.

\item \textit{Real-time processing:} Provide predictable, low-latency on-device execution (no cloud) with a small compute footprint to support repeatable, real-world deployment.

\end{itemize}

The primary \textbf{contributions} of this paper are as follows:

\begin{itemize} 
\item A methodology that integrates two complementary modules: a learning module and a robot action execution, bridging the gap between natural-language parsing and precise robot control.
\item We proposed a custom \textbf{instruction-to-action dataset} and a lightweight \textbf{Instruct2Act} framework that lowers computational overhead while accurately extracting fine-grained sub-actions from human instructions.
\item We introduced a \textbf{RAN} that couples a DATRN with a vision-based environment analyzer to generate smooth, precise manipulation, placement, and interaction within the robot’s workspace.
\end{itemize}
 
The rest of the paper is structured as follows. The proposed methodology is presented in Section II. Section III presents the experimental results and relevant discussions. The work is finally concluded in Section IV.

\section{Proposed Methodology}
\begin{figure*}
    \centering
    \includegraphics[width=\textwidth]{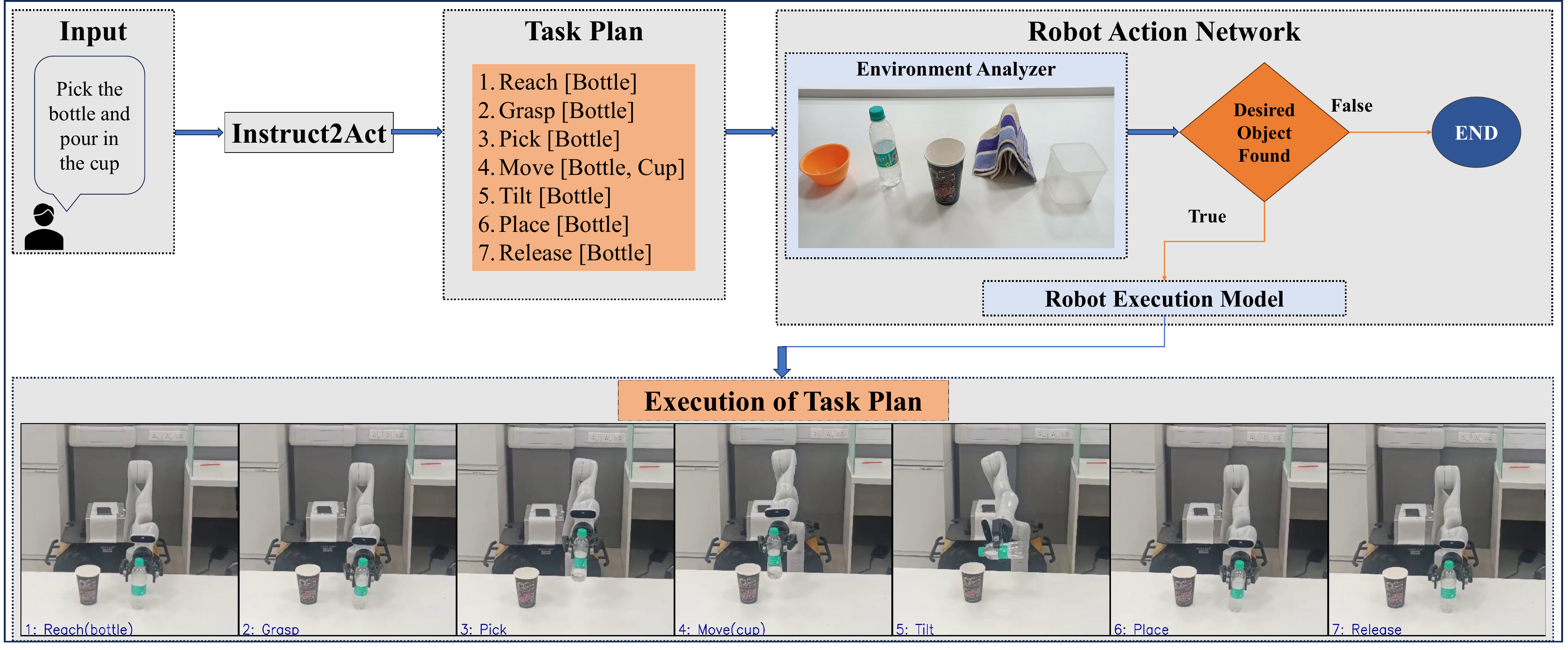}
    \caption{Overall framework: The Instruct2Act provides a task plan, i.e., a sequence of identified sub-actions and the objects from the user's input. The environment analyzer then checks for the target object; if available, the robot execution model executes the sub-actions.}
    \label{fig:fig2}
\end{figure*}

The proposed methodology consists of two modules: (a) Instruct2Act, which translates spoken or written instructions into a sequence of sub-actions that form a task plan for the robot, and (b) RAN, which converts these sub-actions into precise control commands. For spoken input, an offline speech-to-text model (e.g., Whisper) first transcribes the instruction into text, ensuring reliable operation without internet connectivity. Figure~\ref{fig:fig2} illustrates the overall workflow.

\begin{table}[H]
\caption{Sub-actions vocabulary for robot task execution}
\centering
\resizebox{\columnwidth}{!}{%
\begin{tabular}{|l|p{0.50\columnwidth}|}
\hline
\textbf{Actions} & \textbf{Description} \\
\hline
\texttt{reach(\{object\})} & The robot extends its arm toward the object. \\
\hline
\texttt{grasp(\{object\})} & The robot securely grips the object. \\
\hline
\texttt{lift(\{object\})} & The robot lifts the object from its surface. \\
\hline
\texttt{move(\{object\}, \{destination\})} & The robot moves the object to a target destination. \\
\hline
\texttt{tilt(\{object\}, \{container\})} & The robot tilts the object for pouring something into the container. \\
\hline
\texttt{give(\{object\}, to\_person)} & The robot gives the object to a person. \\
\hline
\texttt{release(\{object\})} & The robot releases its grip on the object. \\
\hline
\texttt{place(\{object\}, \{location\})} & The robot places the object at a specified location. \\
\hline
\texttt{wipe(table)} & The robot cleans the table surface. \\
\hline
\texttt{stir(\{container\})} & The robot stirs the container's contents. \\
\hline
\end{tabular}
}
\label{tab:robot_actions}
\end{table}



\subsection{Instruction-to-actions (Instruct2Act)}

The Instruct2Act translates natural-language task descriptions into ordered sequences of sub-actions. 
To capture richer contextual semantics, each instruction is first embedded using the \texttt{BERT-large-uncased} model, which we employ solely for task embedding extraction. The Instruct2Act is based on a bidirectional LSTM with a multi-head attention autoencoder. The objective is to translate natural language task descriptions into sub-action sequences, improving the robot's ability to understand and perform complex tasks. Each step of the approach is discussed below.

\begin{minipage}{0.94\columnwidth}
\centering
\begin{lstlisting}[basicstyle=\ttfamily\footnotesize,
% caption=Action sequence for "Pick the object and place it on the given location.",
caption=Action sequence for ``Pick the object and place it on the given location''.,
label={lst:pick_place}, columns=fullflexible]
(instruction: "Pick the object and place it on the given location";
    actions: (
        "reach({object})";
        "grasp({object})";
        "lift({object})";
        "move({object}, {location})";
        "place({object}, {location})";
        "release({object})"
    )
)
\end{lstlisting}
\end{minipage}

\subsubsection{Data preparation}

We introduce a proprietary dataset in English for fine-grained robotic manipulation comprising 2{,}850 natural-language instructions in total. Of these, 2{,}280 are used for model development with an 80/20 split (1{,}792 train, 448 validation), and a held-out test set of 570 unseen instructions is reserved to assess generalization to novel phrasings and task compositions. Each instruction is paired with a structured sequence of sub-actions and associated objects (see Listing~\ref{lst:pick_place}). The corpus spans multiple task types (pick and place, pick and pour, table cleaning, give, and compositional variants), at least ten distinct workspace objects, and diverse linguistic styles (synonyms, paraphrases, and multi-clause commands). For the primitive action vocabulary used throughout this work, see Table~\ref{tab:robot_actions}, and for example sub-action sequences, see Table~\ref{tab:subtasks_list}.


\begin{table}[!ht]
\centering
\caption{Each task associated with a specific sequence of actions (shown in columns) that serve as labels for identification}
\label{tab:subtasks_list}
\resizebox{\columnwidth}{!}{%
\begin{tabular}{lcccccccccc}
\toprule
\textbf{Task} & 
\textbf{Reach} & 
\textbf{Grasp} & 
\textbf{Lift} & 
\textbf{Move} & 
\textbf{Tilt} & 
\textbf{Place} & 
\textbf{Stir} & 
\textbf{Wipe} & 
\textbf{Release} & 
\textbf{Retract} \\
\midrule
Pick \& Place & 
\checkmark & \checkmark & \checkmark & \checkmark &        & \checkmark &        &        & \checkmark & \checkmark \\

Pick \& Pour & 
\checkmark & \checkmark & \checkmark & \checkmark & \checkmark &           &        &        & \checkmark & \checkmark \\

Stir & 
\checkmark & \checkmark & \checkmark & \checkmark &          &           & \checkmark &        & \checkmark & \checkmark \\

Cleaning & 
\checkmark & \checkmark &           &           &          &           &           & \checkmark & \checkmark & \checkmark \\

Pick \& Give & 
\checkmark & \checkmark & \checkmark & \checkmark &          &           &           &        & \checkmark & \checkmark \\

Pick-Up & 
\checkmark & \checkmark & \checkmark &           &          &           &           &        &           &           \\





\bottomrule
\end{tabular}%
}
\end{table}

\subsubsection{Task embedding extraction}
BERT is used to convert task descriptions into numerical representations that capture their semantic meaning. Specifically, \texttt{bert-large-uncased} model \cite{bert} is used to generate embeddings. The model tokenizes each task description into word pieces and processes them through its transformer layers. This allows for grasping the contextual significance of each token, capturing the nuances and relationships between words with sensitivity to their surrounding context.

\subsubsection{Sub-action sequences} 

To represent sequences of sub-actions as numerical data, each sequence is first padded to a uniform length \( L \). Each sub-action is converted into a one-hot encoded vector, transforming symbolic action sequences into fixed-size numerical tensors for neural network training.


\subsubsection{Learning architecture}
The objective is to design an architecture capable of learning the relationship between task embeddings and sub-action sequences.

    \paragraph{Input layer} The architecture begins with BERT embeddings \(E_i\) derived from task descriptions, where each embedding has a dimension of $d$. These embeddings capture semantic relationships and contextual information from the input text. Formally, for a task description \(T\), the embedding is represented as $    E_i = \text{BERT}(T), $ where $i$ is the task index.


    \paragraph{Bidirectional LSTM (BiLSTM) layer}
    The repeated instruction embeddings denoted as \( R_{i,t} \) for the embedding vector of instruction \( i \) at time step \( t \), are processed by a bidirectional LSTM that reads the sequence in both forward and backward directions. At each time step \( t \), the forward LSTM computes a hidden state \( \overrightarrow{h}_t \) and a cell state \( \overrightarrow{c}_t \), and the backward LSTM computes \( \overleftarrow{h}_t \) and a cell state \( \overleftarrow{c}_t \). These are computed as
    \begin{equation}
    \small \overrightarrow{h}_t = \text{LSTM}(R_{i,t}, \overrightarrow{h}_{t-1}, \overrightarrow{c}_{t-1}) , \overleftarrow{h}_t = \text{LSTM}(R_{i,t}, \overleftarrow{h}_{t+1}, \overleftarrow{c}_{t+1}).
    \end{equation}
    The final BiLSTM output at each time step is the concatenation \( \mathbf{h}_t^{\text{bi}} = [\overrightarrow{h}_t; \overleftarrow{h}_t] \).


    \paragraph{Multi-head attention (MHA) layer}
    The MHA layer enables the architecture to take care of information from different temporal positions within the input sequence, allowing it to capture diverse contextual relationships \cite{MHA}. The hidden representations \( \mathbf{h}^{\text{bi}}_t \) of the BiLSTM layer serve as input to the MHA module.
    
    \begin{itemize}
        \item \textit{Projection into subspaces:}  
        The BiLSTM outputs are linearly projected into \( h \) subspaces to compute multiple sets of queries ($Q$), keys ($K$), and values ($V$) for each attention head,
        \begin{equation}
            Q_i = \mathbf{h_t}^{\text{bi}} W_i^Q
            ,\quad K_i = \mathbf{h_t}^{\text{bi}} W_i^K
            \quad \text{and} \quad
            V_i = \mathbf{h_t}^{\text{bi}} W_i^V,
        \end{equation}
        where \( \mathbf{h_t}^{\text{bi}} \in \mathbb{R}^{L \times d} \) is BiLSTM outputs, and \( W_i^Q, W_i^K, W_i^V \in \mathbb{R}^{d \times d_k} \) are learnable projection matrices for head \( i \).
    
        \item \textit{Attention calculation:}  
        Each head computes scaled dot-product attention as
        \begin{equation}
            \mathrm{attention}(Q_i, K_i, V_i) = \mathrm{softmax}\left( \frac{Q_i K_i^\top}{\sqrt{d_k}} \right) V_i,
        \end{equation}
        where \( d_k \) is the dimensionality of the key vectors. This mechanism assigns dynamic weights to different time steps based on contextual relevance.
    
        \item \textit{Multi-head output:}  
        The outputs \( a_i \) from all heads are concatenated and linearly transformed as
        \begin{equation}
            \mathrm{output} = \mathrm{concat}(a_1, a_2, \ldots, a_h) W^O,
        \end{equation}
        where \( W^O \in \mathbb{R}^{hd_k \times d_{\text{out}}} \) is a learnable projection matrix.
    \end{itemize}
    
    The resulting attention output (\( \mathbb{R}^{L \times d_{out}} \)) is passed through a small feed-forward block, implemented as a bidirectional LSTM. Layer normalization is then applied across the feature dimension. The final sequence representation after this stage is thus \( \mathbb{R}^{L \times d_{out}} \).
\paragraph{Autoencoder layer} 
To encourage the learning of compact and robust representations, we incorporate a lightweight autoencoder after the attention mechanism. This module is trained jointly with the classification head and consists of the following components:
\begin{itemize}
  \item \emph{Bottleneck (encoder):} The output of the preceding BiLSTM feed-forward layer, of shape $\mathbb{R}^{L\times d_{\mathrm{out}}}$, is compressed by a bidirectional LSTM to a bottleneck representation of shape $\mathbb{R}^{L\times d_{z}}$ (with $d_{z} < d_{\mathrm{out}}$). This layer serves as the bottleneck that produces the latent code.
  \item \emph{Reconstruction (decoder):} The latent code is fed to a bidirectional LSTM decoder that reconstructs features back to $\mathbb{R}^{L\times d_{out}}$ (i.e., the decoder output and bottleneck have the same shape).
  \item \emph{Joint training:} We compute an MSE loss between the reconstructed output and the original bottleneck features (both in $\mathbb{R}^{L\times d_{out}}$). This reconstruction loss is weighted and combined with the classification loss during training.
\end{itemize}
This autoencoder block encourages the model to preserve essential task-related information and acts as a regularizer, improving generalization.

\paragraph{Dense output layer}
Finally, a time-distributed dense layer with softmax activation predicts a probability distribution over the set of sub-actions at each time step:
\begin{equation}
    \mathbf{P}_t = \operatorname{softmax}\!\big(W\,\mathbf{d}_t + b\big),\quad t=1,\dots,L,
\end{equation}
where $\mathbf{d}_t \in \mathbb{R}^{d_{out}}$ denotes the decoder output at time step $t$. The weight matrix and bias have shapes $
W \in \mathbb{R}^{C\times d_{out}}, b \in \mathbb{R}^{C}$, so each \( \mathbf{P}_t \in \mathbb{R}^{C} \) represents the predicted class probabilities over \( C \) sub-action categories.

\subsubsection{Model training}
We train the model end-to-end using a combined loss function defined as
\begin{multline}
\mathcal{L} 
= \sum_{t=1}^L \Bigl(-\sum_{i=1}^C y_{t,i}\log p_{t,i}\Bigr) \\
\quad +\; \lambda\, \mathrm{MSE}\bigl(\mathrm{bottleneck},\, \mathrm{reconstruction}\bigr),
\label{eq:loss both}
\end{multline}
where the first term is categorical cross-entropy loss over sub-action labels \( y_{t,i} \), and the second term is mean squared error (MSE) between bottleneck features and their autoencoder reconstructions. The hyperparameter \( \lambda \) controls the trade-off between classification accuracy and reconstruction regularization. 

We optimize this objective using the Adam optimizer, with early stopping based on validation loss. A learning rate scheduler is employed to reduce the learning rate when the validation performance plateaus, promoting stable and efficient convergence.


\subsection{Robot action network (RAN)}
\begin{figure}[h!]
  \centering
  \includegraphics[width=0.98\columnwidth]{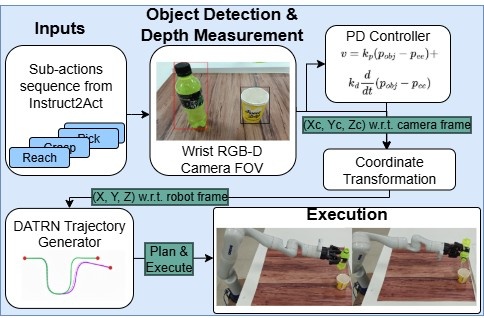}
  \caption{Workflow of RAN.}
  \label{fig:RAN}
\end{figure}

During execution, the RAN plays a crucial role in processing labels received from the recognition model. The RAN executes each sub-actions in sequence to achieve the overall goal, as illustrated in Fig.~\ref{fig:RAN}. It has three components: \textit{(a) optimal trajectory learning, (b) environment analyzer, and (c) robot execution.}

\subsubsection{Optimal trajectory learning}

We model each sub-action trajectory using a dynamic adaptive trajectory radial network (DATRN), which learns smooth, human-like motions from demonstrations. Training data were collected by manually guiding the manipulator from the initial to the desired pose and recording end-effector coordinates; these trajectories serve as the input for DATRN.

The trajectory is modeled using discrete dynamics for reference tracking. The next state ($y(t_{k+1})$) in the trajectory is predicted as
\begin{multline}
  y(t_{k+1}) = y(t_k) + \Delta t\,\Bigl(
    -\,y(t_k)
    + \sum_{i=1}^{N_1} w_{1,i}\,\phi_{1,i}(t_k) \\[-2pt]
    + \sum_{j=1}^{N_2} w_{2,j}\,\phi_{2,j}(t_k)\,u(t_k)
    + \tanh\!\bigl(g - y(t_k)\bigr)
  \Bigr).
  \label{eq:datrn_model}
\end{multline}
Here, \(y(t_k)\) is the state of the trajectory at time \(t_k\), \(\Delta t\) is the integration step, \(g\) the target goal, \(\phi_{k,\ell}(t_k)\) the activation for RBF center \(c_{k,\ell}\), \(w_{k,\ell}\) its associated weight, \(u(t_k)\) an external input, and \(\tanh(g - y(t_k))\) the nonlinear attractor driving stable convergence.

To effectively capture complex motions, the placement of RBF centers is optimized by applying K-Means clustering to the reference trajectories, ensuring centers are concentrated in regions of high variability. These RBFs are Gaussian kernels, with their width $\sigma$ automatically set by the mean distance between neighboring centers to guarantee good coverage. The activation for each RBF at discrete time instants $t_k = p\Delta t$ (where $p = 0,1,\dots,N-1$) is computed as
\begin{equation}
  \phi_{k,\ell}(t_k)
  = \exp\!\biggl(-\frac{(t_k - c_{k,\ell})^2}{2\sigma^2}\biggr),
  \label{eq:gaussian_function}
\end{equation}
for $k \in \{1,2\}$ indexing the two types of RBF network, with $\ell=1,\dots,N_k$ as the center index. Finally, instead of slow iterative training, the entire set of network weights $\mathbf{W}$ is learned efficiently in a single shot using a regularized least-squares (ridge regression) solution,
\begin{equation}
  \label{eq:ridge_solution_condensed}
  \mathbf{W} = (\Phi^\top \Phi + \lambda I)^{-1} \Phi^\top \mathbf{t},
\end{equation}
where \(\Phi\) contains the RBF activations and \(\mathbf{t}\) is the target state derivatives. This single-shot learning approach produces smooth, human-like motion that adapts to task variations without retraining. By ensuring each sub-action follows an optimized trajectory, DATRN enhances overall robotic performance and preserves the essential characteristics of demonstrated movements.

\subsubsection{Environment analyzer}
The environment analyzer uses a custom-trained YOLOv8-n detector 
to localize referenced items in the workspace. A sub-action that specifies an object, the analyzer searches the scene either \textit{(i)} returns the object's target location if detected with sufficient confidence, or \textit{(ii)} if not detected, reports that the object is not found and safely halts the task. If an instruction or sub-action omits the object or destination, the system interactively queries the user for the missing source and goal. Based on human feedback, once the required items are confirmed in view, their poses are passed to the action network for execution; otherwise, the episode is terminated.


\subsubsection{Robot execution model}

The robot execution model focuses on executing tasks based on identified sub-actions. The model operates through a series of well-defined steps:

\paragraph{Object detection and depth measurement}
The robot first processes the RGB image through the YOLOv8-n network to identify the target object. A bounding box is drawn around the detection, and its center pixel coordinates \((u,v)\) are computed. Simultaneously, the depth sensor provides a per-pixel distance map; indexing this map at \((u,v)\) yields the depth of the object \(d\).

\paragraph{Error minimization}
Tasks that require reaching two or more target locations, such as moving a bottle from the initial coordinates to the cup’s pour coordinates, are especially challenging in an eye-in-hand RGB-D setup. As soon as the robot grasps an object, the camera’s view is partially occluded, so the pick and subsequent place or pour decisions must be made in advance. To guarantee precise execution, we employ a proportional-derivative (PD) controller. At each control step, it computes the velocity command,
\[
\mathbf{v}(t)
=
K_{p}\bigl(\mathbf{p}_{\mathrm{obj}}(t) - \mathbf{p}_{\mathrm{ee}}(t)\bigr)
+K_{d}\,\frac{\mathrm{d}}{\mathrm{d}t}\bigl(\mathbf{p}_{\mathrm{obj}}(t) - \mathbf{p}_{\mathrm{ee}}(t)\bigr),
\]
where $\mathbf{p}=[x,y,z]^\mathsf{T}$ and $\mathbf{p}_{\mathrm{ee}}, \,\mathbf{p}_{\mathrm{obj}}$ denote the positions of the effector and the center of the object, respectively. $K_{p}$ and $K_{d}$ are the proportional and derivative gains, respectively. For $\|\mathbf{p}_{\mathrm{obj}} - \mathbf{p}_{\mathrm{ee}}\|<\epsilon$, the arm holds its pose at the desired grasp or pour point, and object center ($x, y,$ and $z$ coordinates) is updated to extract the optimal grasp point subsequently.

\paragraph{Coordinate transformation}
The image-plane 2D coordinates \((u,v)\) of the detected object's bounding box center and its depth measurement \(d\) are back-projected to camera frame 3D coordinates via:
\begin{equation}
X_c = \frac{(u - c_x)}{f_x} \cdot d, \quad
Y_c = \frac{(v - c_y)}{f_y} \cdot d, \quad
Z_c = d,
\end{equation}
where $(c_x, c_y)$ denote the camera principal point and $(f_x, f_y)$ are the focal lengths in pixels. 
The resulting camera-frame coordinates $(X_c, Y_c, Z_c)$ are then transformed into the robot-base frame 
using the homogeneous transformation matrix $\mathbf{T}_{c}^{b}$, obtained from ROS TF2, as
$   [x, y, z, 1]^\top = \mathbf{T}_{c}^{b} \cdot [X_c, Y_c, Z_c, 1]^\top $.

\paragraph{DATRN-based trajectory planning and execution}
The robot frame coordinates obtained \((x,y,z)\) are provided to the DATRN. It generates a smooth, human-like trajectory that guides the manipulator to the target, and the robot executes this trajectory to accurately reach and interact with the object.

\section{Results and Discussion}
The experiments were carried out to evaluate the execution of tasks using the Instruct2Act-based approach. The results offer insight into the effectiveness of the model.

\subsection{Experimental setup}
The setup has hardware and software components.

\subsubsection{Hardware} A Kinova 7-DOF Robot Manipulator is utilized to perform various tasks. Positioned on top of the manipulator is a visual sensor (Intel RealSense Depth Module D410) used during task execution for depth calculation. Model training is performed using a single GPU (Nvidia RTX 4060 with 8GB of VRAM).

\subsubsection{Software} To control and interact with the robot manipulator and camera, we use the robot operating system (ROS) and Kinova libraries \cite{kinova_gen3_user_guide}. We have used the Huggingface Transformers, TensorFlow, and PyTorch frameworks for efficient training.

\subsection{Instruct2Act implementation details}

We instantiate the proposed Instruct2Act model as follows (symbols defined in Table~\ref{tab:notation}). The task text $T$ is encoded with BERT in a vector $E_i\!\in\!\mathbb{R}^{d}$ ($d{=}1024$), which is tiled (or temporally aligned) to length $L$ to form $\{R_{i,t}\in\mathbb{R}^{d}\}_{t=1}^L$. A BiLSTM with $256$ units per direction consumes $R_{i,t}$ and outputs $h^{\mathrm{bi}}\!\in\!\mathbb{R}^{L\times 512}$ . Multi-head self-attention takes $h^{\mathrm{bi}}$ and, for each head $j\!\in\!\{1,\dots,h\}$ with $h{=}8$, projects $h^{\mathrm{bi}}_t$ to $Q_j,K_j,V_j\!\in\!\mathbb{R}^{L\times d_k}$ via $W_j^{Q},W_j^{K},W_j^{V}\!\in\!\mathbb{R}^{512\times d_k}$ with $d_k{=}256$. The head outputs are concatenated ($\mathbb{R}^{L\times (h d_k)}\!=\!\mathbb{R}^{L\times 2048}$) and linearly mapped by $W^{O}\!\in\!\mathbb{R}^{2048\times 512}$ to $\tilde{H}\!\in\!\mathbb{R}^{L\times 512}$. A second BiLSTM (128 units per direction) acts as a lightweight feed-forward block, yielding $H^{\mathrm{ff}}\!\in\!\mathbb{R}^{L\times 256}$, followed by layer normalization. 
The autoencoder then compresses $H^{\mathrm{ff}}$ with a 
BiLSTM (64 units per direction) to the latent sequence $Z\!=\!\{z_t\}_{t=1}^L$, $z_t\!\in\!\mathbb{R}^{128}$, and decodes via a BiLSTM (128 units per direction) to $\hat{B}\!\in\!\mathbb{R}^{L\times 256}$. A time-distributed dense layer with parameters $W\!\in\!\mathbb{R}^{256\times C}$ and $b\!\in\!\mathbb{R}^{C}$ maps $\hat{b}_t$ to class probabilities $\mathbf{P}_t\!=\!\mathrm{softmax}(W\hat{b}_t{+}b)\!\in\!\mathbb{R}^{C}$. Training is end-to-end with Adam (initial LR $1{\times}10^{-4}$), early stopping, and loss $\mathcal{L}\!=\!\sum_{t=1}^{L}\!\big(-\sum_{c=1}^{C} y_{t,c}\log p_{t,c}\big) + \lambda\,\mathrm{MSE}(B,\hat{B})$.

\begin{table}[t]
\centering
\small
\caption{Notations used in implementation}
\label{tab:notation}
\resizebox{0.99\linewidth}{!}{%
\begin{tabular}{@{}ll@{}}
\toprule
Symbol & Meaning \\
\midrule
$T$ & Text (instruction) for \lq task\rq \\
$E_i\!\in\!\mathbb{R}^{d}$ & BERT embedding of task $i$; here $d{=}1024$ \\
$L$ & Sequence length after tiling/alignment \\
$R_{i,t}\!\in\!\mathbb{R}^{d}$ & Input at time $t$ to the first BiLSTM \\
$h^{\mathrm{bi}}\!\in\!\mathbb{R}^{L\times 512}$ & BiLSTM sequence output \\
$h$ & Number of attention heads (here $h{=}8$) \\
$d_k$ & Head key/query/value size (here $d_k{=}256$) \\
$W_j^{Q},W_j^{K},W_j^{V}$ & Projections $\mathbb{R}^{512\times d_k}$ for head $j$ \\
$W^{O}$ & Output projection $\mathbb{R}^{(h d_k)\times 512}$ \\
$H^{\mathrm{ff}}\!\in\!\mathbb{R}^{L\times 256}$ & Post-attention BiLSTM output \\
$B,\hat{B}\!\in\!\mathbb{R}^{L\times 256}$ & Bottleneck and its reconstruction \\
$Z{=}\{z_t\}$, $z_t\!\in\!\mathbb{R}^{128}$ & Latent sequence (per-step latent vector) \\
$C$ & Number of sub-action classes \\
$\mathbf{P}_t\!\in\!\mathbb{R}^{C}$ & Per-step class probability vector \\
$\lambda$ & Weight for the reconstruction term in $\mathcal{L}$ \\
\bottomrule
\end{tabular}}
\end{table}
 \begin{table*}[!ht]
  \centering
  \caption{Overall: Task‑wise evaluation, failure analysis, and performance comparison}
  \label{tab:combined-eval}
  
  \begin{subtable}{\textwidth}
    \centering
    \caption{Task‑wise Evaluation}
    \label{tab:task_eval}
    \begin{tabular}{|l|c|c|c|c|c|}
      \hline
      Task           & Successful Runs & Failed Runs & Success Rate (\%) 
                     & Sub-action Prediction Time (s) & Task Execution Time (s) \\
      \hline
      Pick \& Place  & 18               & 2           & 90               
                     & 2.5-3.8                              & 35-45 \\
      Pick \& Pour   & 18               & 2           & 90               
                     & 2.5-3.8                              & 45-50 \\
      Table Cleaning & 19               & 1           & 95               
                     & 2.5-3.8                              & 60-70 \\
      Pick \& Give   & 17               & 3           & 85               
                     & 2.5-3.8                             & 35-45 \\
      \hline
    \end{tabular}
  \end{subtable}
  \vspace{1em}
  \begin{subtable}{0.49\textwidth}
    \centering
    \caption{Failure analysis (total 8 failures across all tasks)}
    \label{tab:failure_analysis}
    \begin{tabular}{|l|c|c|}
      \hline
      Failure Cause                   & \ Failures & Percentage (\%) \\
      \hline
      Robot Action Network            & 3           & 37.5             \\
      Sub-action Sequence Prediction   & 2           & 25               \\
      System Failure                  & 3           & 37.5             \\
      \hline
    \end{tabular}
  \end{subtable}
  \hfill
  \begin{subtable}{0.49\textwidth}
    \centering
    \caption{Performance comparison with baseline models}
    \label{tab:model_comparison}
    \resizebox{0.99\textwidth}{!}{%
    \begin{tabular}{|l|c|c|c|c|c|}
      \hline
      Model           & Accuracy (\%) & Parameters   & \makecell{Train\\Time (min)}
                      & \makecell{Weighted F1\\Score (\%)} & \makecell{Weighted\\Recall (\%)} \\
      \hline
      LSTM            & 79.21         & 1.32M        & 1.2        
                      & 79                                & 77 \\
      BiLSTM          & 81.92         & 2.63M        & 1.7        
                      & 81                                & 79 \\
      BiLSTM + MHA    & 85.56         & 6.83M        & 3.1        
                      & 89                                & 87 \\
      Instruct2Act        & \textbf{91.5} & \textbf{7.91M} & \textbf{4.7} 
                      & \textbf{91}                       & \textbf{88} \\
      \hline
    \end{tabular}}
  \end{subtable}

\end{table*}

 \subsection{Instruct2Act learning evaluation}
\subsubsection{Training and validation loss}
\begin{figure}[!ht]
  \centering
  \includegraphics[width=\columnwidth]{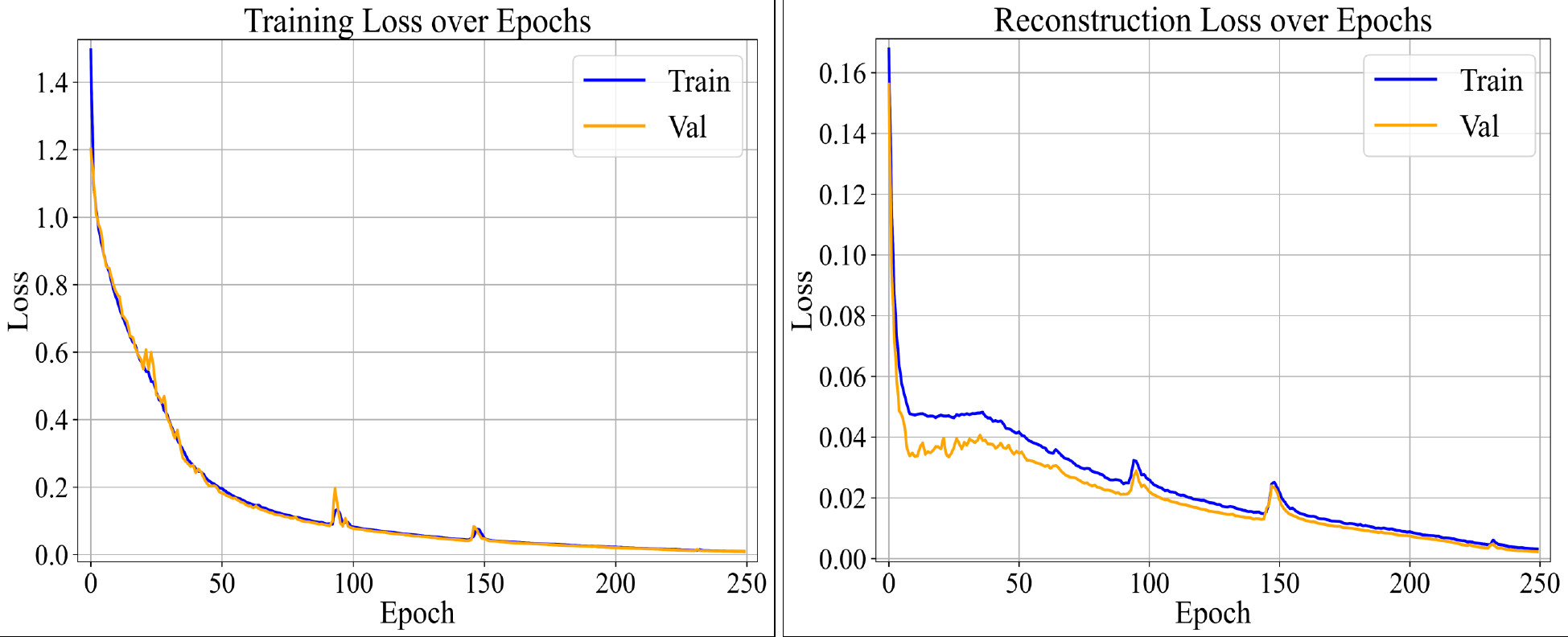}
  \caption{Training and reconstruction loss curves for the model.}
  \label{fig:train_val}
\end{figure}

Instruct2Act was trained on instruction–action pairs using a batch size of 64, an initial learning rate of $1\times10^{-4}$, and early stopping (patience = 5, min. $\delta$ = $1\times10^{-6}$) for up to 300 epochs. As shown in Fig.~\ref{fig:train_val}, both training and validation losses declined smoothly, converging at 0.0091 and 0.0088, respectively. This parallel convergence demonstrates effective learning without overfitting.

Likewise, the reconstruction loss for the autoencoder branch decreased steadily on both the training and validation sets, confirming that the model captures meaningful latent representations while maintaining robust generalization.  

\begin{figure}[!ht]
  \centering
  \includegraphics[width=0.90\columnwidth]{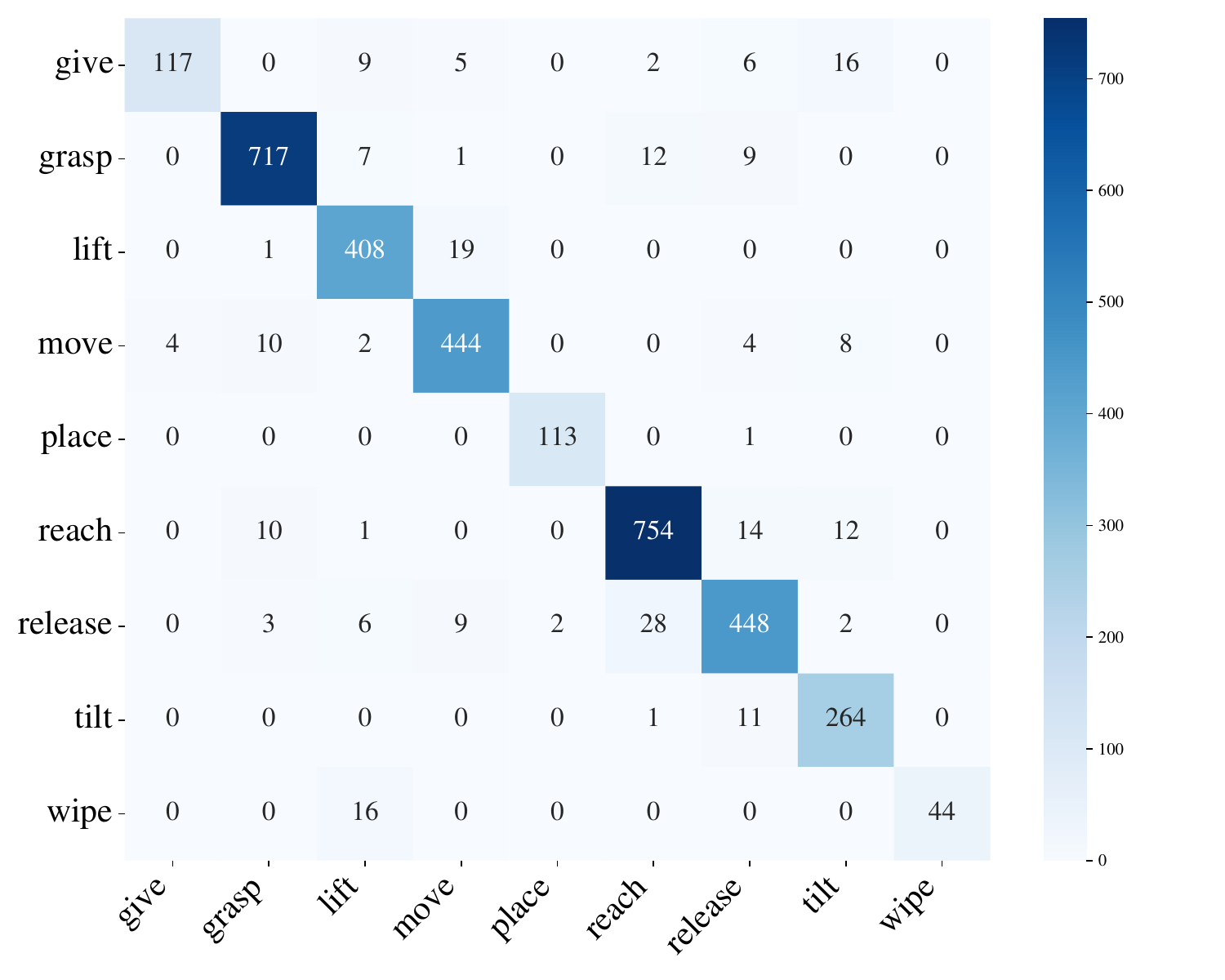}
  \caption{Confusion matrix of Instruct2Act on test set.}
  \label{fig:confusion_matrix}
\end{figure}

\subsubsection{Comparative performance analysis}

To quantify Instruct2Act’s advantages, we benchmarked it against three standard sequence models using the same training and test splits and an identical preprocessing pipeline. Each model was trained end-to-end on ground-truth sub-action labels, with only architecture-specific hyperparameters tuned for optimal performance. Table~\ref{tab:model_comparison} reports the total parameter count, training time, and key classification metrics, while Fig.~\ref{fig:confusion_matrix} displays the confusion matrix for our top-performing model.

Our Instruct2Act model trained with a batch size of 64 and requiring only 6 GB of GPU achieves the highest accuracy (91.5\%), weighted F1 score (91\%), and weighted recall (88\%). This outperforms the BiLSTM + MHA baseline (85.6 \% accuracy, 89\% F1, 87\% recall) and the simpler LSTM variants (82.2\% and 81.9\% accuracy). Although Instruct2Act incurs a modest increase in parameters (7.91 M) and training time (4.7 min), the integration of multi‐head attention and an autoencoder yields a substantial boost in predictive power, affirming its suitability for real‐time robotic sub-action recognition on limited GPU hardware.

\subsection{Trajectory learning with DATRN evaluation}
The RAN leverages DATRN to convert each predicted sub-action and its object/goal pose into an end-effector path. In our deployments, DATRN adds negligible compute overhead relative to perception and control.

\begin{figure}[!ht]
  \centering
  \includegraphics[width=\linewidth,page=1]{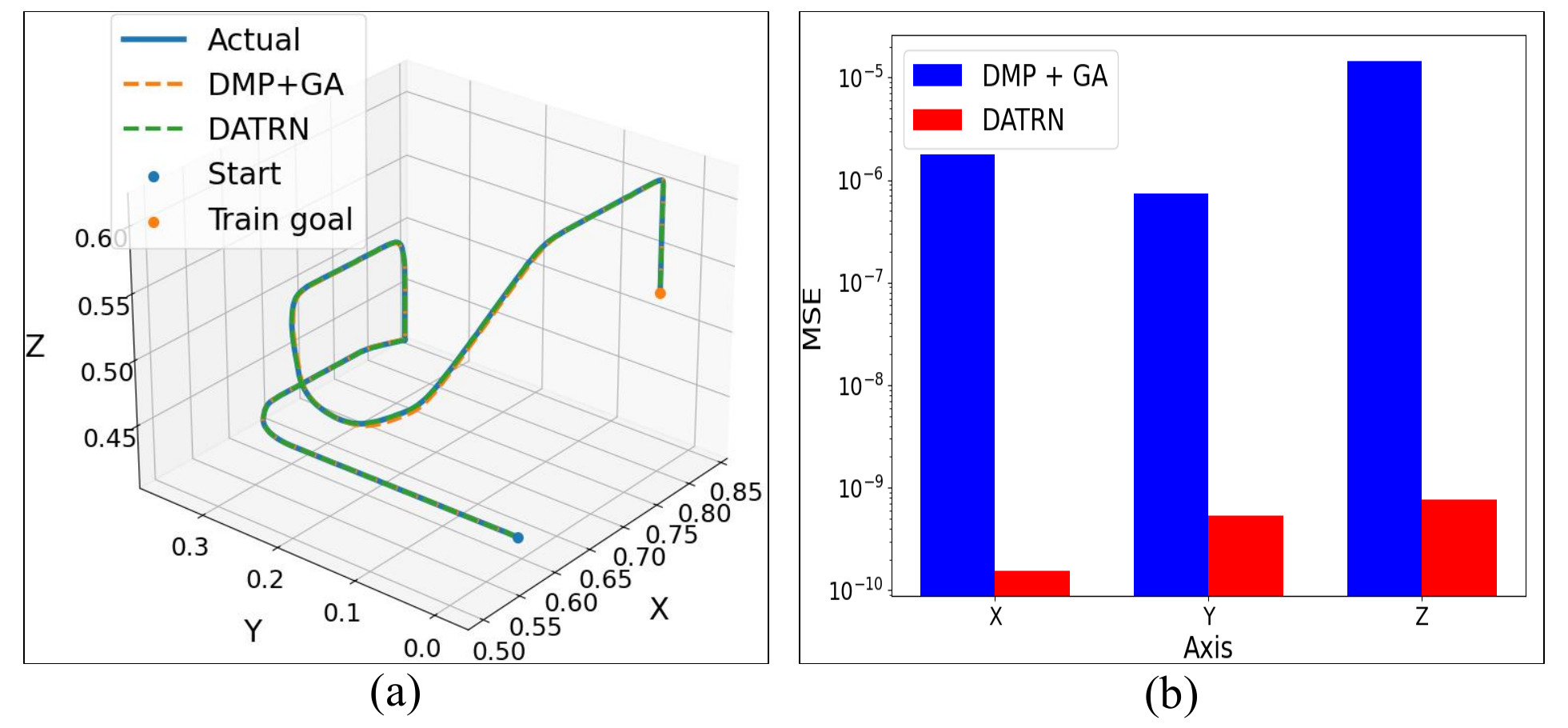}
  \caption{(a) Trajectory prediction using DATRN and DMP+GA and (b) Per-axis fitting error: DATRN and DMP+GA across \(X,Y,Z\).}
  \label{fig:datrn_dmp_mse}
\end{figure}

\begin{figure*}[!ht]
    \centering

    \begin{subfigure}{0.49\textwidth}
        \centering
        \setlength{\fboxsep}{0.5pt}
        \fbox{\includegraphics[width=0.985\linewidth]{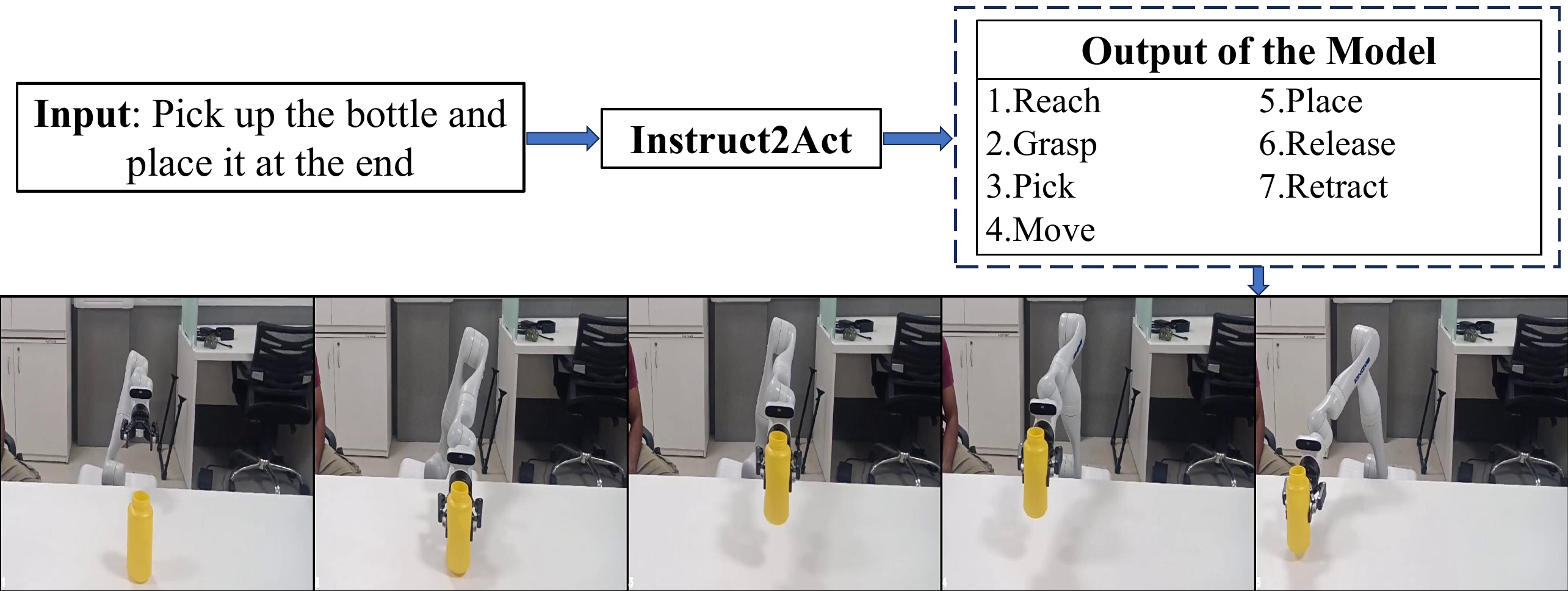}}
        \caption[Pick \& Place]{Pick and place: Environment analyzer detects bottle. The system performs reach, grasp, pick, move, place, release, and retract.}
        \label{fig:pick_and_place}
    \end{subfigure}
    \hfil
    \begin{subfigure}{0.49\textwidth}
        \centering
        \setlength{\fboxsep}{0.5pt}
        \fbox{\includegraphics[width=0.985\linewidth]{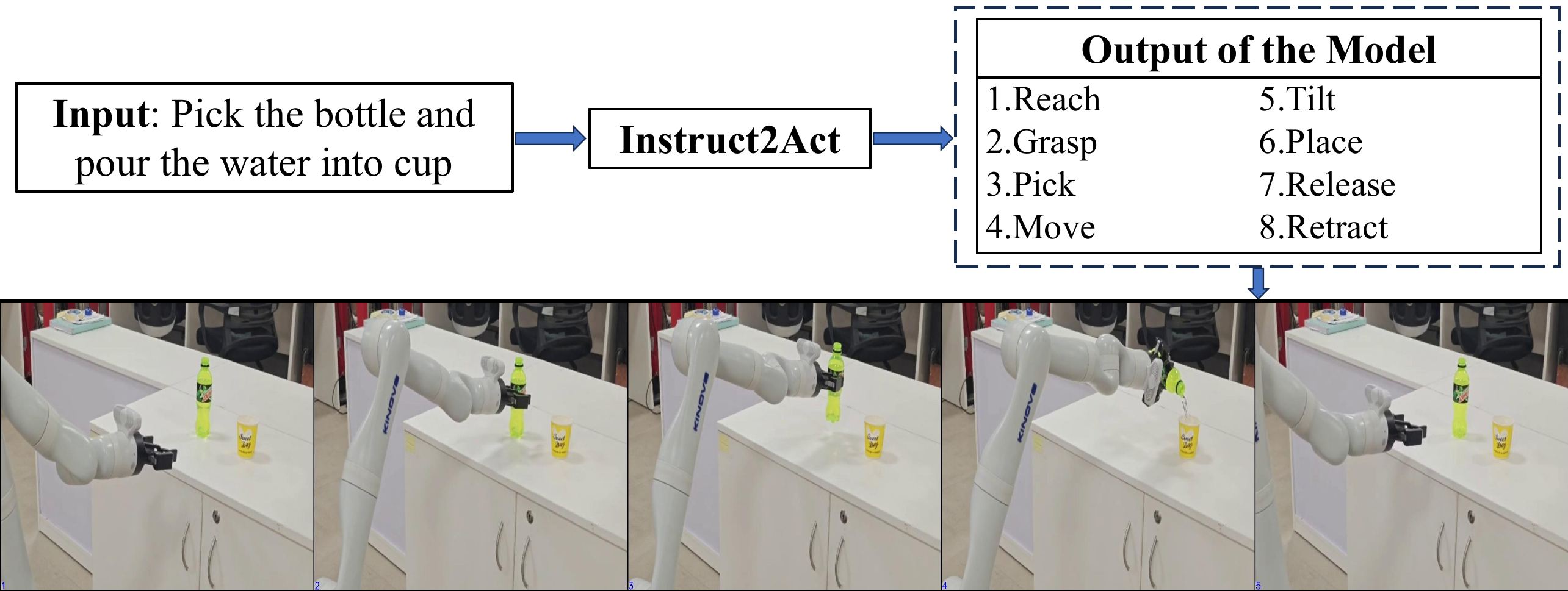}}
        \caption[Pick \& Pour]{Pick and pour: Analyzer locates bottle and glass. The system executes reach, grasp, pick, move, tilt, place, release, and retract.}
        \label{fig:pick_and_pour}
    \end{subfigure}
    \hfil
    \begin{subfigure}{0.49\textwidth}
        \centering
        \setlength{\fboxsep}{0.5pt}
        \fbox{\includegraphics[width=0.985\linewidth]{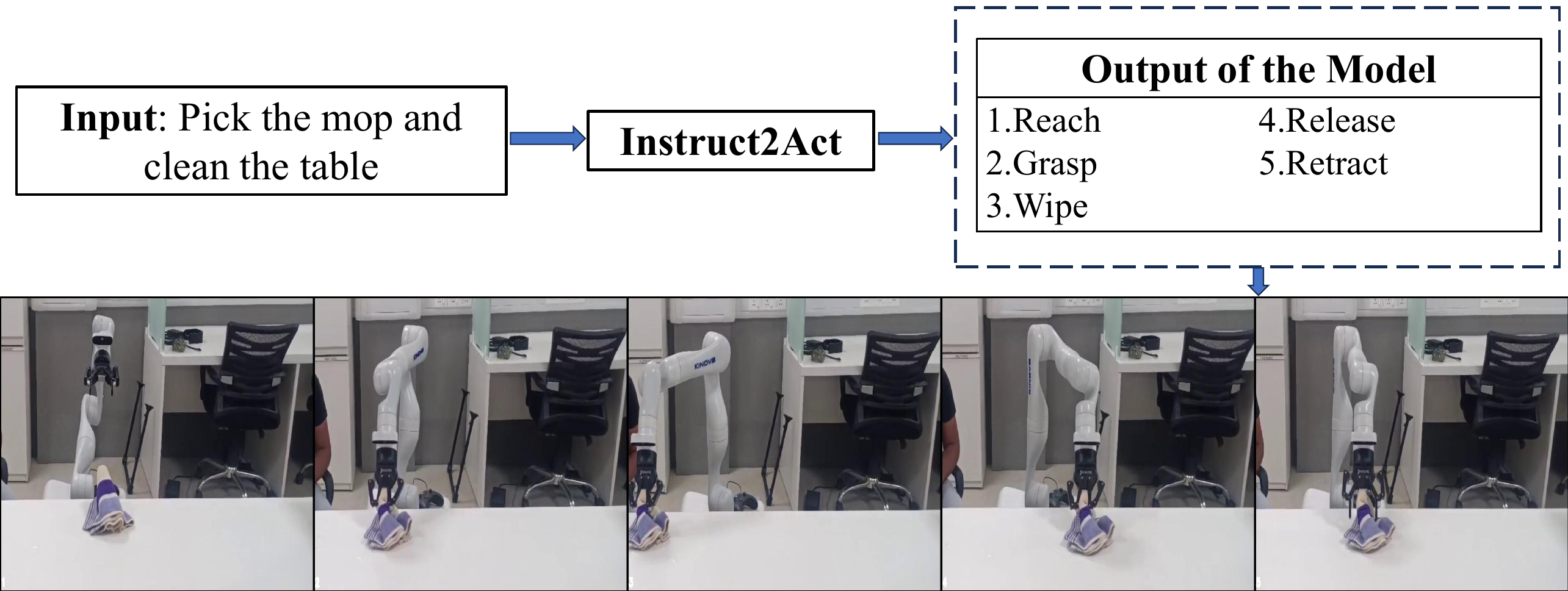}}
        \caption[Table Cleaning]{Table cleaning: Analyzer detects cleaning tool and surface. The robot performs reach, grasp, wipe, release, and retract.}
        \label{fig:wipe}
    \end{subfigure}
    \hfil
    \begin{subfigure}{0.49\textwidth}
        \centering
        \setlength{\fboxsep}{0.5pt}
        \fbox{\includegraphics[width=0.985\linewidth]{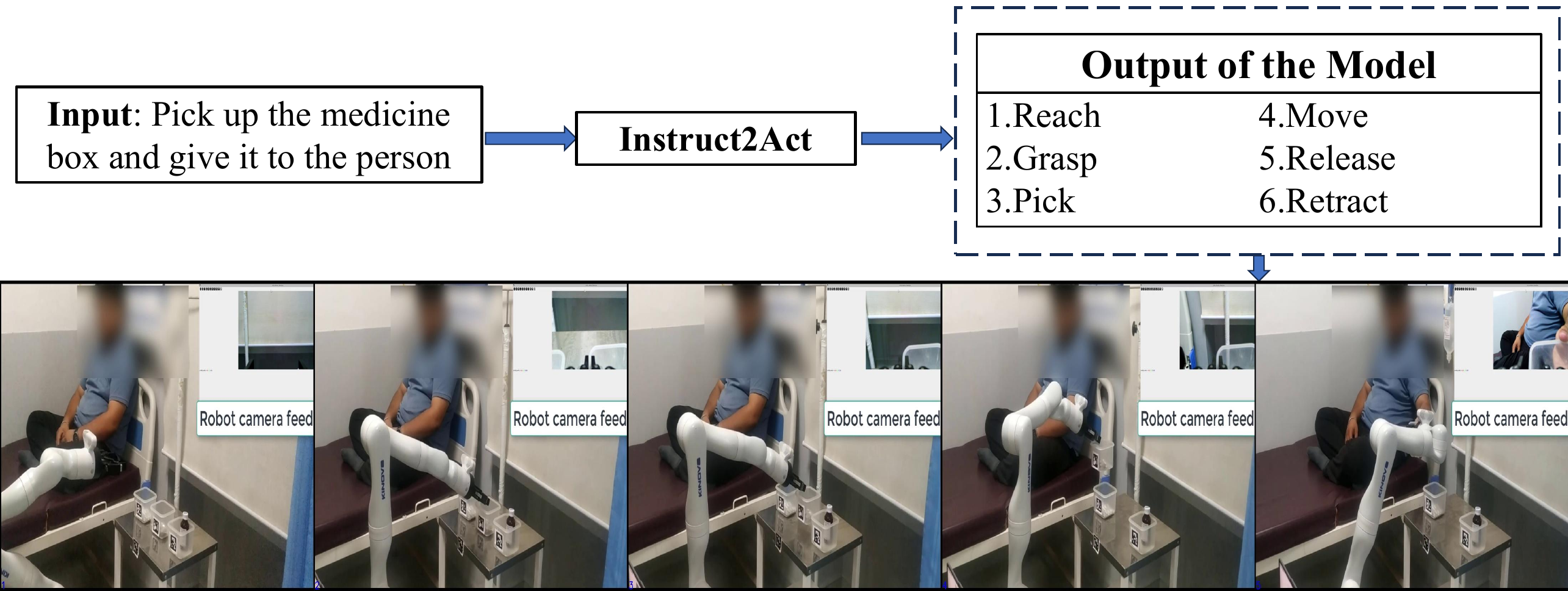}}
        \caption[Pick \& Give]{Pick and give: Analyzer identifies object and recipient. The robot performs reach, grasp, pick, move, release, and retract.}
        \label{fig:pick_and_give}
    \end{subfigure}

    \caption{Robot performing diverse manipulation tasks as instructed in natural language. Each subfigure displays a sub-actions sequence coordinated by the environment analyzer and task-specific controllers.}
    \label{fig:all_tasks}
\end{figure*}

\subsubsection{DATRN vs.\ DMP} We evaluated DATRN against DMP on each task trajectory. For DMP, we used a genetic algorithm (GA) search to avoid manual trial-and-error; this produced good trajectories with improved fitting accuracy. However, in a head-to-head evaluation, DATRN was still better on both fronts: it achieved higher trajectory fidelity and required less prediction time. The path overlays show DATRN closely tracking the demonstration across high-curvature segments and near the goal, whereas DMP+GA remains slightly less accurate. The per-axis MSE confirms DATRN’s advantage (see Fig.~\ref{fig:datrn_dmp_mse}). Across trajectory lengths ranging from approximately  $400$ to $900$ data points, DATRN completes training in only $1.82$--$2.50\,$s. This corresponds to an approximately $95\%$ reduction in training time relative to DMP+GA, which requires $61.11$--$98\,$s for the same tasks.

\begin{figure}[H]
  \centering
  \includegraphics[width=\linewidth,page=1]{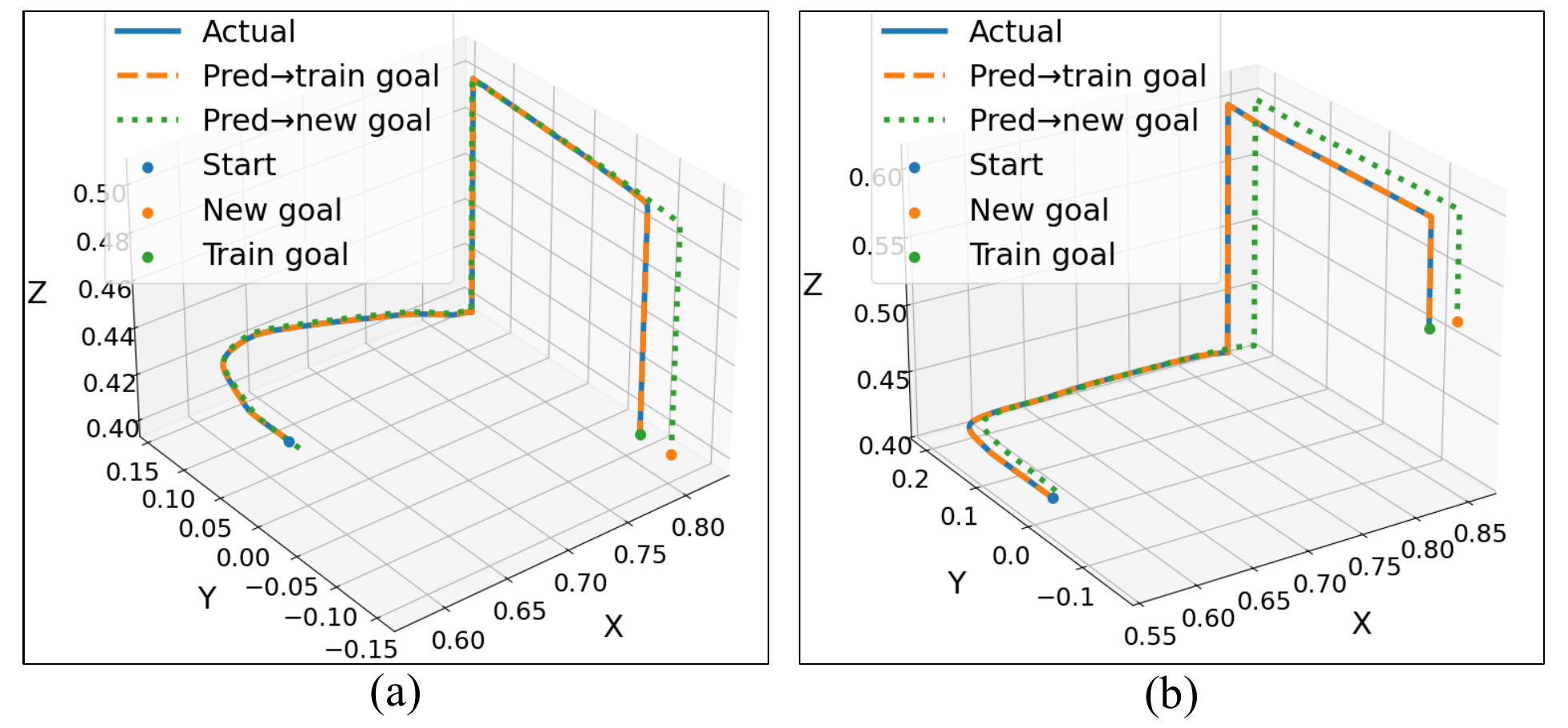}
  \caption{Goal-adaptive trajectory prediction with DATRN. The predicted path adapts online to the target pose.}
  \label{fig:datrn_goal_adapt}
\end{figure}

DATRN’s advantage stems from its structure: (i) it learns weights in closed form via ridge regression without iterative tuning, (ii) it does not require phase variables or time-dependent gain/variance schedules; (iii) it uses fewer, more interpretable hyperparameters (RBF count/centers, ridge coefficient). In practice, this makes DATRN faster to fit, more accurate at goal tracking, and simpler to deploy on our on-device, low-latency setup. Figure~\ref{fig:datrn_goal_adapt} illustrates goal adaptivity in placing: as the goal pose changes, the path reshapes and re-times smoothly, maintaining low endpoint error and avoiding overshoot, which enables reliable execution across various source$\rightarrow$goal configurations. Hence, we adopt DATRN as the default trajectory learner for all experiments and deployments.

\subsection{End-to-End tasks evaluation}

We evaluated the end-to-end performance of our system across four representative manipulation tasks in both controlled laboratory experiments and some trials in healthcare environments. 
Fig.~\ref{fig:pick_and_place} shows the pick and place task, 
Fig.~\ref{fig:pick_and_pour} shows the pick and pour task, 
Fig.~\ref{fig:wipe} shows the table cleaning task, 
and Fig.~\ref{fig:pick_and_give} shows the pick and give task (demonstrated in a hospital environment). 
Each figure demonstrates the robot’s ability to interpret a natural-language instruction, detect the target object, and execute the corresponding sub-action sequence. 

Although this paper presents four primary tasks, our system also supports complex instructions ranging from single actions to multi-step sequences (e.g., pick and pour followed by give, or pick and place followed by surface cleaning).

\subsubsection{Task performance and execution analysis}
We evaluated our system on four distinct manipulation tasks, conducting 20 trials for each. As detailed in Table~\ref{tab:task_eval}, the system demonstrated high reliability, achieving a 95\% success rate for table cleaning and 90\% for both pick and place and pick and pour. The success rate for the more complex, interactive Pick and Give task was 85\%. Critically, the sub-action prediction time from a given instruction was consistently under 3.8 seconds, affirming the system's suitability for real-time applications. The total task execution time for each task is shown in Table~\ref{tab:task_eval}, which correlates with the physical complexity and duration of the required actions for each task.

\subsubsection{Failure analysis}

Across 80 trials, our system achieved a 90\% success rate. We analyzed the 8 recorded failures across all these tasks. As detailed in Table~\ref{tab:failure_analysis}, issues occurred at the execution level: 37.5\% from the physical robot action network; these failures were caused by perception errors like object detection inaccuracy, motion singularities, and grasp slippage. Another 37.5\% from general system failures (e.g., system lags, robot hangs). The remaining 25\% of failures originated at the planning stage, caused by incorrect sub-action sequence prediction.

\section{Conclusion}

This paper presents a two-stage pipeline that converts free-form instructions into a robot task plan, specifically a sequence of sub-actions. The proposed framework integrates the Instruct2Act model, which captures the temporal and contextual structure of human instructions, with a RAN that includes a DATRN for trajectory planning and an environmental analyzer for interpreting relevant environmental data. The Instruct2Act model predicts sub-actions from human instructions, and the RAN translates these predictions into executable motions for diverse tasks. Our system demonstrates reliable and resource-efficient end-to-end execution, achieving an overall 90\% accuracy across four manipulation tasks in controlled laboratory experiments and preliminary trials in real-world healthcare environments under single eye-in-hand sensing.

Although Instruct2Act does well on short and medium-horizon instructions, it has trouble with very long or highly compositional commands, usually accurately predicting early sub-actions but missing later steps because of trouble preserving long-range dependencies. To address these limitations, we plan to expand the dataset with more diverse, long-horizon instructions across varied objects and settings, and to explore lightweight transformer-based architectures that preserve low computational overhead while improving sub-action prediction for complex, multi-step activities.

\bibliographystyle{IEEEtran}
\bibliography{ref}

@online{kinova_gen3_user_guide,
  title = {Kinova Gen3 Ultra lightweight robot User Guide},
  author = {Kinova Robotics},
  year = {2020},
  url = {https://www.kinovarobotics.com/uploads/User-Guide-Gen3-R07.pdf},
}

@article{bert,
  author    = {Jacob Devlin and
               Ming{-}Wei Chang and
               Kenton Lee and
               Kristina Toutanova},
  title     = {{BERT:} Pre-training of Deep Bidirectional Transformers for Language
               Understanding},
  journal   = {CoRR},
  volume    = {abs/1810.04805},
  year      = {2018},
  url       = {http://arxiv.org/abs/1810.04805},
  archivePrefix = {arXiv},
  eprint    = {1810.04805},
  bibsource = {dblp computer science bibliography, https://dblp.org}
}

@article{MHA,
  title={Multi-head attention: Collaborate instead of concatenate},
  author={Cordonnier, Jean-Baptiste and Loukas, Andreas and Jaggi, Martin},
  journal={arXiv preprint arXiv:2006.16362},
  year={2020}
}

@INPROCEEDINGS{1,
  author={Zhou, Haotian and Lin, Yunhan and Yan, Longwu and Zhu, Jihong and Min, Huasong},
  booktitle={2024 IEEE International Conference on Robotics and Automation (ICRA)}, 
  title={LLM-BT: Performing Robotic Adaptive Tasks based on Large Language Models and Behavior Trees}, 
  year={2024},
  volume={},
  number={},
  pages={16655-16661},
  doi={10.1109/ICRA57147.2024.10610183}}

@ARTICLE{2,
  author={Vemprala, Sai H. and Bonatti, Rogerio and Bucker, Arthur and Kapoor, Ashish},
  journal={IEEE Access}, 
  title={ChatGPT for Robotics: Design Principles and Model Abilities}, 
  year={2024},
  volume={12},
  number={},
  pages={55682-55696},
  doi={10.1109/ACCESS.2024.3387941}}

@ARTICLE{4,
  author={Liu, Haokun and Zhu, Yaonan and Kato, Kenji and Tsukahara, Atsushi and Kondo, Izumi and Aoyama, Tadayoshi and Hasegawa, Yasuhisa},
  journal={IEEE Robotics and Automation Letters}, 
  title={Enhancing the LLM-Based Robot Manipulation Through Human-Robot Collaboration}, 
  year={2024},
  volume={9},
  number={8},
  pages={6904-6911},
  doi={10.1109/LRA.2024.3415931}}

@misc{7,
      title={Large Language Models Can Self-Improve}, 
      author={Jiaxin Huang and Shixiang Shane Gu and Le Hou and Yuexin Wu and Xuezhi Wang and Hongkun Yu and Jiawei Han},
      year={2022},
      eprint={2210.11610},
      archivePrefix={arXiv},
      primaryClass={cs.CL},
      url={https://arxiv.org/abs/2210.11610}, 
}

@misc{8,
      title={Deployment of NLP and LLM Techniques to Control Mobile Robots at the Edge: A Case Study Using GPT-4-Turbo and LLaMA 2}, 
      author={Pascal Sikorski and Leendert Schrader and Kaleb Yu and Lucy Billadeau and Jinka Meenakshi and Naveena Mutharasan and Flavio Esposito and Hadi AliAkbarpour and Madi Babaiasl},
      year={2024},
      eprint={2405.17670},
      archivePrefix={arXiv},
      primaryClass={cs.RO},
      url={https://arxiv.org/abs/2405.17670}, 
}

@ARTICLE{RobotGPT,
  author={Jin, Yixiang and Li, Dingzhe and A, Yong and Shi, Jun and Hao, Peng and Sun, Fuchun and Zhang, Jianwei and Fang, Bin},
  journal={IEEE Robotics and Automation Letters}, 
  title={RobotGPT: Robot Manipulation Learning From ChatGPT}, 
  year={2024},
  volume={9},
  number={3},
  pages={2543-2550},
  doi={10.1109/LRA.2024.3357432}}

@INPROCEEDINGS{VLM,
  author={Gao, Jensen and Sarkar, Bidipta and Xia, Fei and Xiao, Ted and Wu, Jiajun and Ichter, Brian and Majumdar, Anirudha and Sadigh, Dorsa},
  booktitle={2024 IEEE International Conference on Robotics and Automation (ICRA)}, 
  title={Physically Grounded Vision-Language Models for Robotic Manipulation}, 
  year={2024},
  volume={},
  number={},
  pages={12462-12469},
  doi={10.1109/ICRA57147.2024.10610090}}

@article{Intro_RAL_1,
  title={Understanding natural language instructions for fetching daily objects using gan-based multimodal target--source classification},
  author={Magassouba, Aly and Sugiura, Komei and Quoc, Anh Trinh and Kawai, Hisashi},
  journal={IEEE Robotics and Automation Letters},
  volume={4},
  number={4},
  pages={3884--3891},
  year={2019},
  publisher={IEEE}
}

@article{NLP_robot_survey,
  title={Language-conditioned learning for robotic manipulation: A survey},
  author={Zhou, Hongkuan and Yao, Xiangtong and Meng, Yuan and Sun, Siming and Bing, Zhenshan and Huang, Kai and Knoll, Alois},
  journal={arXiv preprint arXiv:2312.10807},
  year={2023}
}

@ARTICLE{tinyvla,
  author={Wen, Junjie and Zhu, Yichen and Li, Jinming and Zhu, Minjie and Tang, Zhibin and Wu, Kun and Xu, Zhiyuan and Liu, Ning and Cheng, Ran and Shen, Chaomin and Peng, Yaxin and Feng, Feifei and Tang, Jian},
  journal={IEEE Robotics and Automation Letters}, 
  title={TinyVLA: Toward Fast, Data-Efficient Vision-Language-Action Models for Robotic Manipulation}, 
  year={2025},
  volume={10},
  number={4},
  pages={3988-3995},
  doi={10.1109/LRA.2025.3544909}}

@misc{openvla,
    title={OpenVLA: An Open-Source Vision-Language-Action Model},
    author={Moo Jin Kim and Karl Pertsch and Siddharth Karamcheti and Ted Xiao and Ashwin Balakrishna and Suraj Nair and Rafael Rafailov and Ethan Foster and Grace Lam and Pannag Sanketi and Quan Vuong and Thomas Kollar and Benjamin Burchfiel and Russ Tedrake and Dorsa Sadigh and Sergey Levine and Percy Liang and Chelsea Finn},
    year={2024},
    eprint={2406.09246},
    archivePrefix={arXiv},
    primaryClass={cs.RO}
}

@misc{smolvla,
    title={SmolVLA: A Vision-Language-Action Model for Affordable and Efficient Robotics},
    author={Mustafa Shukor and Dana Aubakirova and Francesco Capuano and Pepijn Kooijmans and Steven Palma and Adil Zouitine and Michel Aractingi and Caroline Pascal and Martino Russi and Andres Marafioti and Simon Alibert and Matthieu Cord and Thomas Wolf and Remi Cadene},
    year={2025},
    eprint={2506.01844},
    archivePrefix={arXiv},
    primaryClass={cs.LG}
}

@article{Saveriano,
   title={Dynamic movement primitives in robotics: A tutorial survey},
   volume={42},
   ISSN={1741-3176},
   url={http://dx.doi.org/10.1177/02783649231201196},
   DOI={10.1177/02783649231201196},
   number={13},
   journal={The International Journal of Robotics Research},
   publisher={SAGE Publications},
   author={Saveriano, Matteo and Abu-Dakka, Fares J and Kramberger, Aljaž and Peternel, Luka},
   year={2023},
   month=sep, pages={1133–1184} }

@article{hosiptals,
title = {Human-robot interactions in autonomous hospital transports},
journal = {Robotics and Autonomous Systems},
volume = {179},
pages = {104755},
year = {2024},
issn = {0921-8890},
doi = {https://doi.org/10.1016/j.robot.2024.104755},
url = {https://www.sciencedirect.com/science/article/pii/S0921889024001398},
author = {Andreas Zachariae and Frederik Plahl and Yucheng Tang and Ilshat Mamaev and Björn Hein and Christian Wurll},
}

\end{document}